\documentclass[10pt,twocolumn,letterpaper]{article}

\usepackage{wacv}
\usepackage[accsupp]{axessibility} 
\usepackage{times}
\usepackage{epsfig}
\usepackage{graphicx}
\usepackage{amsmath}
\usepackage{amssymb}
\usepackage[table]{xcolor}
\usepackage[pagebackref=true,breaklinks=true,colorlinks,bookmarks=false,citecolor=blue,linkcolor=blue]{hyperref}

\def\wacvPaperID{1032} 

\wacvfinalcopy 

\ifwacvfinal
\fi

\ifwacvfinal
\usepackage[breaklinks=true,bookmarks=false]{hyperref}
\else
\usepackage[pagebackref=true,breaklinks=true,colorlinks,bookmarks=false]{hyperref}
\fi

\begin{document}

\title{Multi-Head Deep Metric Learning Using Global and Local Representations}

\author{Mohammad K.~Ebrahimpour, Gang Qian, and Allison Beach \\
ObjectVideo Labs, Inc., 8281 Greensboro Dr., Tysons, VA 22102 \\
\small{mkebrahimpour@gmail.com, \{gqian, abeach\}@objectvideo.com}
}


\maketitle
\ifwacvfinal
\thispagestyle{empty}
\fi

\begin{abstract}
Deep Metric Learning (DML) models often require strong local and global representations, however, effective integration of local and global features in DML model training is a challenge. DML models are often trained with specific loss functions, including pairwise-based and proxy-based losses. The pairwise-based loss functions leverage rich semantic relations among data points, however, they often suffer from slow convergence during DML model training. On the other hand, the proxy-based loss functions often lead to significant speedups in convergence during training, while the rich relations among data points are often not fully explored by the proxy-based losses. In this paper, we propose a novel DML approach to address these challenges. The proposed DML approach makes use of a hybrid loss by integrating the pairwise-based and the proxy-based loss functions to leverage rich data-to-data relations as well as fast convergence. Furthermore, the proposed DML approach utilizes both global and local features to obtain rich representations in DML model training. Finally, we also use the second-order attention for feature enhancement to improve accurate and efficient retrieval. In our experiments, we extensively evaluated the proposed DML approach on four public benchmarks, and the experimental results demonstrate that the proposed method achieved state-of-the-art performance on all benchmarks.
\end{abstract}
%
\begin{figure}
    \centering
    \begin{tabular}{cc}
  \rotatebox{90}{\hspace{13ex} R@1}&\includegraphics[width = 0.8 \linewidth]{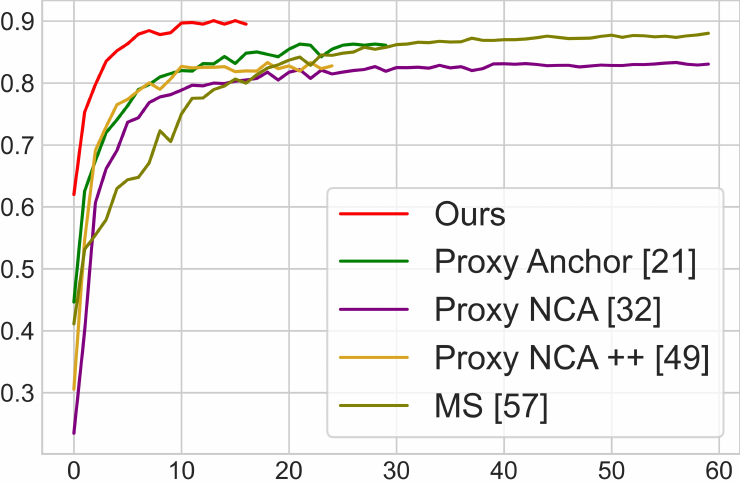} \\
  & Epochs \\
    \end{tabular}
    \caption{Accuracy in Recall@1 versus epochs on the Cars-196~\cite{cars196} dataset. Note that all methods were trained on a single Quadro p5000 GPU with a batch size of 100. Our method achieves the highest accuracy while converging at the same order as the proxy-based baselines in terms of the number of epochs.}
    \label{f:convergence}
\end{figure}
\section{Introduction}
\label{s:intro}
\vspace{-0.2cm}
Learning semantically meaningful representations has been a vital step in numerous computer vision applications such as representation learning~\cite{rep-learning1,rep-learning2},  content-based visual retrieval~\cite{kim-content,proxy-anchor,proxy-NCA,proxy-nca++}, person or vehicle re-identification~\cite{persion-re-id1,persion-re-id2,car-re-id}, and face verification~\cite{face1,face2}. Deep Convolutional Neural Networks (CNNs) have proven repeatedly their effectiveness in the large spectrum of applications~\cite{ebrahimpour_attention, ebrahimpour_audio,ebrahimpour_sensitivity,semantic-segmentation,language,ebrahimpour_wacv,ebrahimpour_ijcnn} including Deep Metric Learning (DML). The neural networks in DML are trained to map the data to a lower-dimensional embedding space in which similar data (data in the same class) are pulled together and the dissimilar data (data in different classes) are pushed away~\cite{proxy-nca++,ms}. For such an embedding space, rich representations and special loss functions are inevitable. 

High image retrieval often requires global and local representations~\cite{global-local-features}. The global features~\cite{global1, global2}, or ``global descriptors" compactly summarize the contents of an image. Often global descriptors are taken from the deepest layers in CNNs; therefore, they only involve the most abstract information, and the vital identifiers such as geometry and spatial information are lost. On the other hand, local features~\cite{local1, local2}, involve information about the geometry and spatial information of the input image. Generally speaking, global features lead to better recall, while local features are essential in better precision~\cite{global-local-features}. A typical retrieval system setup takes advantage of both global and local features in its final embeddings to obtain the best of both worlds. 

Recently, self-attention or Second-Order Attention (SOA) in feature space has received a significant attraction~\cite{soa-application1, soa-application2,solar,sosnet,soa-application3}. The SOA can be considered as a spatial enhancement technique that reflects the correlation among spatial locations and enhances the highly correlated parts of the feature map. Although recent deep-learning-based global descriptors provide effective ways to aggregate features into a compact global vector, they have not explored the correlations of low-level and high-level features within feature maps simultaneously.
\begin{figure}[t]
    \centering
\includegraphics[width = 0.9 \linewidth]{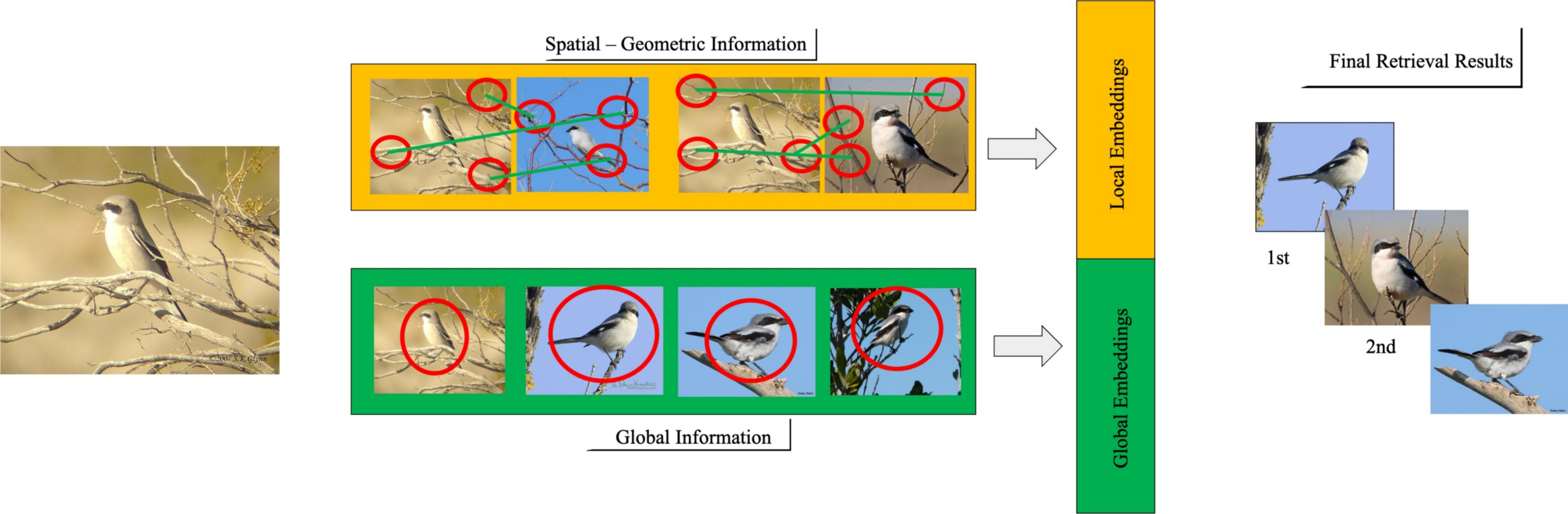} 
    \caption{Our proposed deep metric learning architecture with local and global features. Our model jointly extracts deep local and global features. Both of these features will be further enhanced spatially by an SOA mechanism.}
    \label{f:concept}
\end{figure}
The other vital factor in DML is the loss function. The loss functions are essential to provide a powerful supervisory signal based on the problem objectives~\cite{proxy-anchor,ms}. The loss functions in the DML problems are classified into pairwise-based~\cite{ms,triplet,triplet1,npair,npair1}, and proxy-based~\cite{proxy-NCA,proxy-anchor,proxy-nca++} models. The pairwise-based losses are built based on comparing the pairwise distances between data in the batch. While the pairwise-based losses provide a strong supervisory signal for training the model by considering \emph{data-to-data} relations~\cite{npair,ms}, they suffer from sample mining and slow convergence~\cite{proxy-anchor}. 

The proxy-based losses address the above issues by introducing a limited number of proxies~\cite{proxy-NCA,proxy-anchor,proxy-nca++}. A proxy is a representative of a subset of training data (for instance, a proxy per class) and learned along with network parameters. Since the number of proxies is substantially smaller than the data-points, the proxy-based models benefit from faster convergence rates than the pairwise-based losses. Note the proxy-based models are associated with \emph{data-to-proxy} relations and they miss the rich supervisory information of data-to-data relations.

In this paper, we propose a multi-head network 
that benefits from the fast convergence of the proxy-based loss functions and rich data-to-data relation of the pairwise-based models. Although we are using a hybrid of both proxy-based and pairwise-based loss functions in our multi-head network, our approach does not introduce any hyper-parameter tuning for tuple sampling. Our framework also involves an SOA mechanism to exploit the correlation between features at different spatial locations to further enhance the deep local and global features. Also, we combine both global and local descriptors to produce the final descriptor that holds the content information as well as geometry and spatial information to efficiently select the most similar images. With the above advantages, our proposed method achieves state-of-the-art performance in terms of Recall@1 and quickly converges as exhibited in Figure~\ref{f:convergence}.

The contribution of this paper unfolds as follows:~(a)~We propose a multi-head network that takes advantage of both pairwise-based and proxy-based methods; it leverages rich data-to-data relations and enables fast and reliable convergence.~(b)~We explore the SOA for further enhancement of both local and global features based on higher-order information.~(c)~We demonstrate the impact of using local and global descriptors, proxy-based and pairwise-based, SOA, and the embedding dimensions via a thorough ablation study on their effects.~(d)~An embedding neural network trained with our approach achieves state-of-the-art performance on four publicly available benchmarks for metric learning~\cite{cars196,cub,sop,inshop}.
\begin{figure*}[t]
    \centering
\begin{tabular}{c}
\includegraphics[width = 0.82 \linewidth]{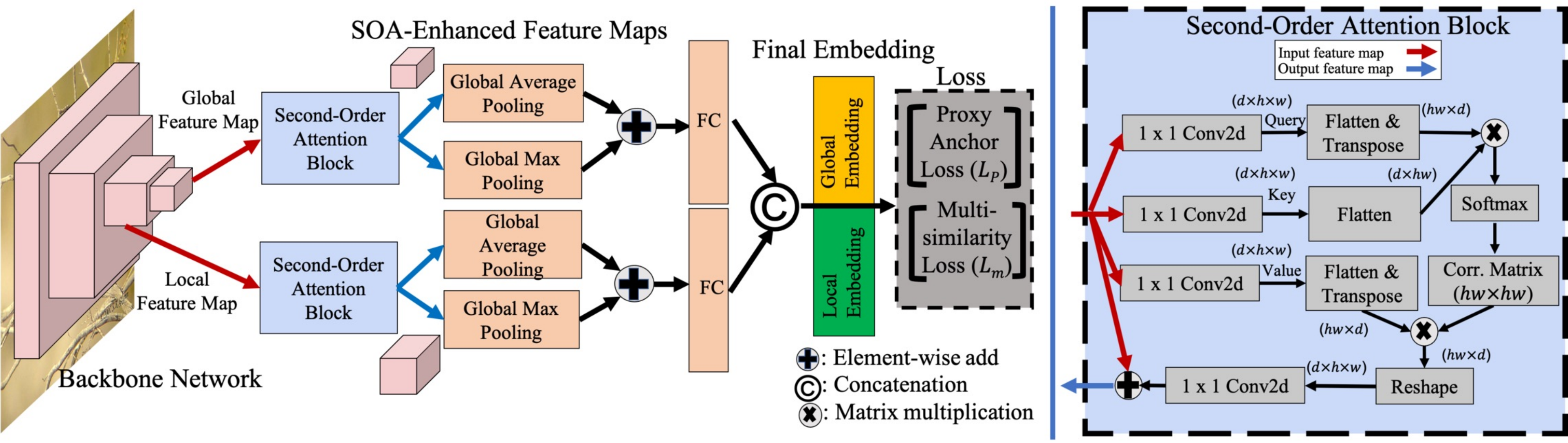} \\
\end{tabular}
    \caption{Our proposed multi-head metric learning architecture with joint local and global features. The local and global features are jointly extracted from the backbone and they are sent to the SOA block (the blue block) for further enhancement. Then, we apply pooling layers on top of the refined and re-weighted features. The final embedding involves the concatenation of local and global representations that are used for retrieval to efficiently select the most similar images based on both local and global identifiers simultaneously. Finally, a hybrid loss involving a proxy-based and a pairwise based is applied to the final embedding.}
    \label{f:proposed_method}
\end{figure*}
\section{Related Work}
\vspace{-0.2cm}
In this section, we categorize the DML approaches based on their use of descriptors and loss functions into two broad categories, then we review relevant papers in each category.
\subsection{Loss Functions}
\vspace{-0.2cm}
Loss functions in DML can be divided into two groups, pairwise-based and proxy-based.

\vspace{0.18cm}
\noindent\textbf{Pairwise-based Losses.} Contrastive loss~\cite{contrastive,contrastive1} and Triplet loss~\cite{triplet,triplet1} are influential examples of loss functions for pairwise-based DML.
Contrastive loss takes a pair of embedding vectors as input and aims to push them apart if they are of different classes or pull them together if they are of the same class. Triplet loss considers a data point as an anchor. Each anchor is associated with a positive (an embedding with an identical class label to the anchor) and a negative data point (an embedding with different class labels) and involves the distance of the anchor-positive pair to be smaller than that of the anchor-negative pair in the embedding space. 

One potential issue with pairwise-based models is that a large number of tuples have a limited contribution to the learning algorithm and sometimes even diminish the quality of the learned embedding space~\cite{sampling}. To address this issue, most pairwise-based losses~\cite{npair,lifted-structure,ranked-list} employ hard sample mining techniques~\cite{sampling,smart-mining}. However, these techniques involve tuning hyper-parameters and consequently increases the risk of over-fitting. Pairwise-based losses are rich in data-to-data relations. However, the number of tuples increases polynomially with regard to the number of training data, resulting in prohibitive complexity and significantly slow convergence~\cite{proxy-anchor}. 

\vspace{0.18cm}
\noindent\textbf{Proxy-based Losses.} Proxy-based metric learning endeavors to address the complexity and slow convergence issue of the pairwise-based losses. The proxy-based methods require a small set of proxies to capture the global structure of an embedding space and assign each data point to relevant proxies instead of the other data points during training. Since the number of proxies is significantly smaller than the training data, the training complexity reduces substantially. For instance, Proxy-NCA~\cite{proxy-NCA} loss assigns a single proxy to each class and associates data points to each proxy and encourages positive pairs to be close together and negative pairs to be far apart. Proxy-NCA++~\cite{proxy-nca++} is an extension of the Proxy-NCA and aims to enhance the limitations of the Proxy-NCA in terms of temperature factor and pooling layer.

SoftTriple loss~\cite{softtriple}, inspired by the Proxy-NCA yet assigns multiple proxies to each class instead of one to improve the likelihood of capturing the intra-class variance. Proxy-Anchor loss~\cite{proxy-anchor} assigns a proxy to each class and treats each proxy as an anchor and assigns positive and negative pairs to each anchor. Although introducing proxies in proxy-based losses significantly improves the convergence in model training, it has an inherent limitation of data-to-proxy relation instead of data-to-data relation that results in limited supervisory information. Our multi-head network overcomes this limitation by proposing a hybrid approach involves in both pairwise-based and proxy-based methods to benefit data-to-data relations as well as high convergence rates.
\subsection{Descriptors}
\vspace{-0.2cm}
DML algorithms can be divided into three groups based on their use of descriptors: local descriptors, global descriptors, and joint local and global descriptors.

\vspace{0.18cm}
\noindent\textbf{Local Descriptors.} Hand-engineered features such as SIFT~\cite{local2} and SURF~\cite{local1} have been widely used and adopted for retrieval systems especially before the deep learning era. The key advantage of local features over global ones for image retrieval is their capacity to perform spatial matching, often by utilizing RANSAC~\cite{ransac}. Due to the efficiency of local features, recently, several deep learning-based local features have been proposed~\cite{lift,deep-local,local-descriptor,lfnet}.

\vspace{0.18cm}
\noindent\textbf{Global Descriptors.} Global descriptors are often involved the most abstract information about the input, leading to high-performance image retrieval. Before the deep learning era, most global descriptors were obtained using the combination of local descriptors~\cite{combined_local1,combined_local2}. However, recently most high-performing global features are obtained based on CNNs~\cite{neural-code,netvlad,end-2-end}.

\vspace{0.18cm}
\noindent\textbf{Joint Local and Global Descriptors.}
Global descriptors are essential for high recalls yet local descriptors are necessary for better precision; therefore, researchers develop hybrid methods to take advantage of both descriptors. For instance, Taira~\etal~\cite{inloc} used NetVLAD~\cite{netvlad} to extract global features for candidate pose retrieval, followed by
dense local feature matching using feature maps from the same network for indoor localization. Simeoni~\etal’s DSM~\cite{local-features-words} detected key points in activation maps from global feature models. Activation channels are interpreted as visual words, to propose correspondences between a pair of images. Cao~\etal ~\cite{global-local-features} extracted global and local features from the same network. They utilize the global descriptors to retrieve the most similar images and then re-rank the retrieved images by local descriptors to increase the precision.
\section{Proposed Algorithm}
\vspace{-0.2cm}
Our proposed algorithm involves two essential components: Refined deep local and global representations along with a multi-head loss function that enables the data-to-data relation as well as fast convergence. Our model design is illustrated in Figure~\ref{f:concept}.
\subsection{Deep Global and Local Representations}
\vspace{-0.2cm}
We propose to leverage hierarchical representations from a CNN to represent different types of descriptors. While deep layers are associated with the most abstract representations and representing higher-level features, the intermediate layers are more informative in terms of local representations and lower-level features. 

Given an image, from the backbone we obtain two feature maps: $f_l\in\mathbb{R}^{ H_l\times W_l \times C_l}$ and $f_g\in\mathbb{R}^{H_g \times W_g \times C_g}$ , representing local ($l$) and global ($g$) feature maps where $H, W, C$ indicate the height, width, and number of channels respectively. For off-the-shelf convolutional networks, $H_g \leq H_l$,
$W_g \leq W_l$, and $C_g \geq C_l$; indicating that deeper layers have larger number of channels and spatially smaller feature maps.
\subsection{Second-Order Attention (SOA)}
\vspace{-0.2cm}
Let $(i_I , j_I )$ in the input image ($I$) correspond to location $(i, j)$ in feature map $f$. To incorporate higher-order spatial information into the feature map, we adopt the second-order attention block~\cite{soa-application1,solar}. A computational flow of the SOA concept is shown in Figure~\ref{f:proposed_method}. For each feature map, we produce two projections of feature map $f$ named \emph{query} $q$, and \emph{key} $k$, each obtained through $1 \times 1$ 2d-convolutions with possible reduction of number of channels ($d$). Then, by
flattening both tensors, we obtain both $q$ and $k$ in $\mathbb{R}^{d \times hw}$. The second-order attention map $A$ is computed as follows:
\begin{equation}
a = \text{softmax}(\zeta q^Tk), 
\label{e:soa_softmax}
\end{equation}
where $\zeta$ is a scaling factor and $a \in \mathbb{R}^{hw \times hw}$, indicating the correlation of each $f_{i,j}$ to the whole map $f$. A third projection of $f$ and \emph{value} $v$ is then obtained by $ 1\times 1$ 2d-convolution, and after flattening, results in $\mathbb{R}^{hw \times d}$ shape. Then, the second-order attention map $f^{soa}$ is obtained from linear combination of the first-order features $f$ and the second-order attention map:
\begin{equation}
    f^{soa} = f + \phi(a \times v),
\end{equation}
where $\phi$ is yet another $1 \times 1$ convolution
to manage the effect of the obtained attention map. Thus, a new feature $f^{soa}_{i,j}$ in the second-order map $f^{soa} \in \mathbb{R}^{h \times w \times d}$, is a function of features from all locations in $f$:
\begin{equation}
    f^{soa}_{i,j} = h(a_{ij} \odot f),
\end{equation}
where $h$ denotes the combination of all convolutional operations within the non-local block. 
\subsection{Pooling}
\vspace{-0.2cm}
To aggregate deep activations in both global and local features, we adopt the combination of Global Max Pooling (GMP) and Global Average Pooling (GAP) as follows:
\begin{equation}
\begin{split}
   f =     
   \frac{1}{W \times H} \sum_{i\in W, j \in H} f^{soa}
   + \text{max}_{(i \in W, j\in H)} f^{soa}
\end{split}
\end{equation}
After the aggregation, we whiten the aggregated representation for both refined local and global representations; we integrate this into our model with two separated fully-connected layers. The fully connected layer associated with enhanced local representations $F_l \in \mathbb{R}^{C_{f^{soa}_l} \times \frac{D}{2}}$ , with
learned bias $b_{f^{soa}_l} \in \mathbb{R}^{ C_{f^{soa}_l}}$, which $C_{f^{soa}_l}$ indicates the number of channels in the $f^{soa}_l$ and $\frac{D}{2}$ is the dimension of the local embedding space. Similarly, we have a fully connected layer associated with enhanced global representations $F_g \in \mathbb{R}^{C_{f^{soa}_g} \times \frac{D}{2}}$ , with
learned bias $b_{f^{soa}_g} \in \mathbb{R}^{ C_{f^{soa}_g}}$, which $C_{f^{soa}_g}$ indicates the number of channels in the $f^{soa}_g$ and $\frac{D}{2}$ is the dimension of the global embedding space.

After computing $F_l\in \mathbb{R}^{D/2}$ and $F_g \in \mathbb{R}^{D/2}$, the final embedding $F \in \mathbb{R}^{D}$ computes by concatenation of the $F_l$ and $F_g$.
\begin{table*}[t]
\caption{Recall@K ($\%$) on the Cars-196~\cite{cars196} and CUB-200-2011~\cite{cub} datasets. Superscripts indicate embedding sizes. Backbone networks of the models are denoted by abbreviations: G–GoogleNet~\cite{g}, BN–Inception with batch normalization~\cite{batch}, R50–ResNet50~\cite{resnet}. For each group of methods, the best performance is bolded and the second best is underlined.}
\centering
\begin{tabular}{l|c||c c c c|c c c c}
\hline
Algorithms & BackBone & \multicolumn{4}{c|}{Cars-196}&\multicolumn{4}{c}{CUB-200-2011} \\

& &R@1 & R@2 & R@4 & R@8 & R@1 & R@2 & R@4 & R@8 \\
\hline
Clustering$^{64}$~\cite{clustering}& BN&58.1 &70.6& 80.3& 87.8&48.2& 61.4& 71.8& 81.9 \\ 
Proxy-NCA$^{64}~\cite{proxy-NCA}$ & BN & 73.2 &82.4 &86.4 &87.8 & 49.2 &61.9 &67.9 &72.4 \\
Smart Mining$^{64}~\cite{smart-mining}$ & G & 64.7 &76.2& 84.2& 90.2&49.8& 62.3& 74.1& 83.3\\
MS$^{64}$~\cite{ms}& BN &78.6& 86.6& 91.8& 95.4&60.1& 71.9& 81.2& 88.5 \\
Proxy-Anchor$^{64}$~\cite{proxy-anchor} & R50 & 78.8& \underline{87.0}& \underline{92.2}& \textbf{95.5}& \underline{61.7}& \underline{73.0}& \underline{81.8}& \underline{88.8} \\ 
 \rowcolor{lightgray}Ours$^{64}$&R50&\textbf{81.1}&\textbf{88.1}&\textbf{92.3}&\underline{95.3}&\textbf{63.1}&\textbf{74.6}&\textbf{83.2}&\textbf{89.4}\\
\hline
Margin$^{128}$~\cite{sampling} & R50 & \underline{79.6}& \underline{86.5}&\underline{91.9}& \underline{95.1}&\underline{63.6}&\underline{74.4}&\underline{83.1}&\underline{90.0} \\
\rowcolor{lightgray} Ours$^{128}$& R50 & \textbf{84.9}&\textbf{90.6}&\textbf{94.0}&\textbf{96.6}&\textbf{66.6}&\textbf{77.1}&\textbf{85.2}&\textbf{91.3}\\
\hline
HDC$^{384}$~\cite{hdc} & G & 73.7& 83.2& 89.5& 93.8&53.6& 65.7& 77.0& 85.6 \\  
A-BIER$^{512}~\cite{a-bier}$&G&82.0& 89.0& 93.2& 96.1&57.5& 68.7& 78.3& 86.2\\
ABE$^{512}$~\cite{attention-ensemble} &G& 85.2& 90.5& 94.0& 96.1& 60.6& 71.5& 79.8& 87.4 \\
HTL$^{512}$ & BN& 81.4& 88.0& 92.7& 95.7&57.1& 68.8& 78.7& 86.5 \\
RLL-H$^{512}$~\cite{ranked-list} & BN& 74.0& 83.6& 90.1& 94.1& 57.4& 69.7& 79.2& 86.9\\
MS$^{512}$~\cite{ms} & R50 & 84.1& 90.4& 94.0& 96.5& 65.7& 77.0& 86.3& 91.2\\
SoftTriple$^{512}$~\cite{softtriple}&BN&84.5& 90.7& 94.5& 96.9&65.4& 76.4& 84.5& 90.4 \\
Proxy-Anchor$^{512}$~\cite{proxy-anchor}&BN&86.1& 91.7& 95.0& 97.3&68.4& 79.2& 86.8& 91.6\\
ProxyNCA++$^{512}~\cite{proxy-nca++}$&R50& 86.5& 92.5& 95.7& 97.7&69.0&79.8&\underline{87.3}&\textbf{92.7} \\
Proxy-Anchor$^{512}$~\cite{proxy-anchor}&R50&\underline{87.7} &\underline{92.9}& \underline{95.8}& \underline{97.9}&\underline{69.7}& \underline{80.0}& 87.0& \underline{92.4} \\
 \rowcolor{lightgray} Ours$^{512}$& R50 & \textbf{90.1}&\textbf{94.2}&\textbf{96.4}&\textbf{98.1}&\textbf{70.6}&\textbf{80.9}&\textbf{88.0}&92.3\\
\hline
\end{tabular}
\label{t:cars_cub}
\end{table*}
\subsection{A Hybrid Loss Function}
\vspace{-0.2cm}
Our loss is designed to overcome the limitation of both proxy-based and pairwise-based models by introducing a hybrid loss involving the proxy-anchor~\cite{proxy-anchor} loss from the proxy-based category and the MS~\cite{ms} loss from the pairwise-based class.

\vspace{0.18cm} 
\noindent\textbf{Proxy-based Loss.} Proxy-anchor loss~\cite{proxy-anchor} assigns a proxy to each class. Proxy-anchor approach considers each proxy as an anchor and associate it with entire data in a batch to find positive and negative samples. The proxy-anchor loss defined as follows:
\begin{equation}
\begin{split}
    \ell_p(X) = \frac{1}{|P^{+}|} \sum_{p \in P^+} \text{log}(1+\sum_{x \in X_p^{+}} \text{exp}(-\alpha (s(x,p)-\delta))) \\
    + \frac{1}{|P|} \sum_{p \in P} \text{log}(1+\sum_{x \in X_p^{-}} 
    \text{exp}(\alpha (s(x,p)+\delta))),
\end{split}
\label{e:proxy_loss}
\end{equation}
where $\delta > 0$ is a margin, $\alpha > 0$ is a scaling factor, $P$ is the set of all proxies, $s(.,.)$ measures the similarity among its arguments ,and $P^{+}$ indicates the set of positive proxies of data in the batch. Also, for each proxy $p$, a batch of embedding vectors $X$ is divided into the set of positive $X^{+}_p$ and negative $X^{-}_p = X - X^{+}_p$ embedding vectors.

By utilizing the proxy-anchor loss we incorporate data-to-proxy relations as well as fast convergence. For incorporating the data-to-data relation, we integrate the MS loss.

\vspace{0.18cm}
\noindent\textbf{Pairwise-based Loss.} We employ the MS loss~\cite{ms} as a pairwise-based loss since it considers the self, negative, and positive similarities. Self-similarity ensures that the instances belonging to a positive class remains closer to the anchor than the instances associated with negative classes. The positive similarity exclusively deals with positive pairs. $\sigma$ represents the similarity margin that controls the closeness of positive pairs by heavily penalizing those pairs whose cosine similarities are less or equal to $\sigma$. The negative similarity ensures that negative samples have similarity with the anchor as low
as possible. The MS loss function is formulated as follows:
\begin{equation}
\begin{split}
    \ell_m(X) = \frac{1}{m} \sum_{i=1}^{m} (\frac{1}{\gamma} \text{log}(1 + \sum_{k \in P_i} \text{exp}( -\gamma (S_{i,k}-\sigma))) \\
    + \frac{1}{\beta} \text{log}(1 + \sum_{k \in N_i} \text{exp}(\beta (S_{i,k}+\sigma))),
\end{split}
\label{e:ms_loss}
\end{equation}
where $\gamma$, $\beta$, and $\sigma$ are hyper-parameters. $m$ is the number of samples, $P_i$,$N_i$ represents positive and negative samples, and $S_{i,j}$ denotes the pairwise similarity between $x_i$ and $x_j$.

\vspace{0.18cm}
\noindent\textbf{Our Objective Function.}
Our hybrid objective function is a combination of proxy-anchor and MS losses balanced by normalization factor $\lambda$:
\begin{equation}
    \mathcal{L} = \ell_m + \lambda \ell_p
\label{e:loss}    
\end{equation}
\section{Experimental Results}
\label{s:experiments}
\vspace{-0.2cm}
\begin{table}[t]
\caption{Recall@K ($\%$) on the SOP. Superscripts indicates embedding sizes. For each group of methods, the best performance is bolded and the second best is underlined.}
\centering
\begin{tabular}{l|c c c c}
\hline
Recall@K &1 & 10 & 100 & 1000\\
\hline
Clustering$^{64}$~\cite{clustering}& 67.0 & 83.7& 93.2& -\\
Proxy-NCA$^{64}$~\cite{proxy-NCA}& 73.7 & - & - & - \\
MS$^{64}$ &74.1& 87.8& 94.7& 98.2\\
SoftTriple$^{64}$~\cite{softtriple}&76.3& \underline{89.1}& \underline{95.3}& - \\
Proxy-Anchor$^{64}$~\cite{proxy-anchor}&\underline{76.5}& 89.0& 95.1& \underline{98.2}\\
\rowcolor{lightgray} Ours$^{64}$&\textbf{77.3}&\textbf{89.5}&\textbf{95.4}&\textbf{98.3}\\
\hline
Margin$^{128}$~\cite{sampling}&\underline{72.7}& \underline{86.2}&\underline{93.8}& \underline{98.0}\\
 \rowcolor{lightgray} Ours$^{128}$&\textbf{79.1}&\textbf{90.6}&\textbf{95.8}&\textbf{98.5}\\
 \hline
HDC$^{384}$~\cite{hdc}&69.5& 84.4& 92.8& 97.7\\
A-BIER$^{512}$~\cite{a-bier}& 74.2& 86.9& 94.0& 97.8\\
ABE$^{512}$~\cite{attention-ensemble}& 76.3 &88.4& 94.8& 98.2\\
HTL$^{512}$~\cite{htl}&74.8& 88.3& 94.8& 98.4\\
RLL-H$^{512}$~\cite{ranked-list}& 76.1& 89.1& 95.4& -\\
MS$^{512}$~\cite{ms}&78.2& 90.5& 96.0& 98.7\\
SoftTriple$^{512}$~\cite{softtriple}&78.3& 90.3& 95.9& -\\
Proxy-Anchor$^{512}$~\cite{proxy-anchor}& 79.1& 90.8& 96.2& 98.7\\
Proxy-NCA++$^{512}$~\cite{proxy-nca++}&\underline{80.7}& \textbf{92.0}& \textbf{96.7}& \textbf{98.9}\\
 \rowcolor{lightgray} Ours$^{512}$&\textbf{81.7}&\textbf{92.0}&\underline{96.6}&\underline{98.8}\\
\hline
\end{tabular}
\label{t:SOP}
\end{table}
\begin{table}[t]
\caption{Recall@K ($\%$) on the In-Shop. Superscripts indicates embedding sizes. For each group of methods, the best performance is bolded and the second best is underlined.}
\centering
\begin{tabular}{l|c c c c}
\hline
Recall@K &1 & 10 & 20 & 40\\
\hline
HDC$^{384}$~\cite{hdc} & 62.1& 84.9& 89.0& 92.3\\
HTL$^{128}$~\cite{htl}& 80.9& 94.3& 95.8& 97.4 \\
MS$^{128}$~\cite{ms}& 88.0& 97.2& 98.1& 98.7 \\
Proxy-Anchor$^{128}$~\cite{proxy-anchor}& \underline{90.8}& \textbf{97.9}& \underline{98.5}& \textbf{99.0} \\
 \rowcolor{lightgray} Ours$^{128}$&\textbf{90.9}&\textbf{97.9}&\textbf{98.6}&\underline{98.9}\\
 \hline
FashionNet$^{4096}$~\cite{inshop}& 53.0& 73.0& 76.0& 79.0 \\
A-BIER$^{512}$~\cite{a-bier}& 83.1& 95.1& 96.9& 97.8\\
ABE$^{512}$~\cite{attention-ensemble}&87.3& 96.7& 97.9& 98.5 \\
MS$^{512}$~\cite{ms}& 89.7& 97.9& 98.5& 99.1 \\
Proxy-Anchor$^{512}$~\cite{proxy-anchor}& \underline{91.5}& \underline{98.1}& \underline{98.8}& \underline{99.1}\\
ProxyNCA++$^{512}$~\cite{proxy-nca++}& 90.4& \underline{98.1}& \underline{98.8}&\textbf{99.2}\\
 \rowcolor{lightgray} Ours$^{512}$&\textbf{93.1}&\textbf{98.3}&\textbf{99.0}&\textbf{99.2}\\
\hline
\end{tabular}
\label{t:Inshop}
\end{table}
In this section, our method is compared with current state-of-the-art methods on four public benchmark datasets employed for deep metric learning~\cite{cars196,cub,sop,inshop}. We also perform a thorough investigation of local and global features, the SOA, the MS loss and proxy-anchor loss, and embedding dimensionality to study their effects on the proposed method.
\subsection{Dataset}
\vspace{-0.2cm}
We evaluated our model on the CUB-200-2011~\cite{cub}, Cars-196~\cite{cars196}, Stanford Online Product (SOP)~\cite{sop} and In-Shop Clothes Retrieval (In-Shop)~\cite{inshop} datasets. For CUB-200-2011, we set aside 5,864 images of its first 100 classes as a
training set and 5,924 images of the other classes as a test set.
For Cars-196, 8,054 images of its first 98 classes are set aside as a training set and 8,131 images of the other classes are used as a 
test set. For SOP, we follow the standard dataset split
in~\cite{lifted-structure,proxy-anchor,proxy-nca++} using 59,551 images of 11,318 classes as a training set
and the remaining 60,502 images as a test set. Also for
In-Shop dataset, we follow the setting in~\cite{proxy-anchor} to use 25,882 images
of the first 3,997 classes as a training set and 28,760 images of
the remaining classes for test set; the test set is further partitioned into a query and gallery sets with 14,218 images of 3,985 classes and  12,612 images of 3,985 classes, respectively.
For all datasets, we set aside $20\%$ of the training set as a validation set for hyper-parameter tuning~\cite{reality-check}.
\subsection{Implementation Setup}
\vspace{-0.2cm}
\noindent\textbf{Backbone:} For a fair comparison~\cite{reality-check} to recent works, the Resnet50~\cite{resnet} pre-trained on ImageNet classification~\cite{imagenet} is adopted as our backbone network.

\vspace{0.18cm}
\noindent\textbf{Global and Local Features:} For all experiments on all datasets, we obtained the global features from \text{resnet50-conv5-x}$\in \mathbb{R}^{7 \times 7 \times 2084}$ layer and local features are extracted from \text{resnet50-conv4-x}$\in \mathbb{R}^{14 \times 14 \times 1024}$ .

\vspace{0.2cm}
\noindent\textbf{Training:} In all of the experiments,  AdamW algorithm~\cite{adamw} has been adopted and used as our optimizer. AdamW has the same update step as Adam~\cite{adam} with separate decays of weights. In all experiments, our model is trained only for 20 epochs and the initial learning rate is fixed to $10^{-4}$. Note that the learning rate for proxies is set to $10^{-2}$ for faster convergence.

\vspace{0.18cm}
\noindent\textbf{Proxy Setting:} We assign a single proxy to each class as suggested in Proxy-NCA~\cite{proxy-NCA} and Proxy-Anchor~\cite{proxy-anchor}. The proxies are initialized with a normal distribution.

\vspace{0.18cm}
\noindent\textbf{Input Setting:} To have a fair comparison with state-of-the-art methods, the input is re-scaled to $256 \times 256$ and then center-cropped to $224 \times 224$. We used random cropping and horizontal flipping during training as the data augmentation strategy, as suggested in~\cite{proxy-anchor,proxy-NCA}. During the test, the images are only center-cropped. The default size of cropped images is fixed to $224 \times 224$~\cite{reality-check}.

\vspace{0.18cm}
\noindent\textbf{Hyperparameter Setting:} $\zeta$ in Eq.~\ref{e:soa_softmax} is set to $1$. $\alpha$ and $\delta$ in Eq.~\ref{e:proxy_loss} is set to $32$ and $10^{-1}$ respectively. $\gamma$, $\beta$, and $\sigma$ in Eq.~\ref{e:ms_loss} are set to $2, 50, 1$. Finally, $\lambda$ in Eq.~\ref{e:loss} is set to $3 \times 10^{-2}$, for all experiments. 
\subsection{Comparison to Other Methods}
 \vspace{-0.2cm}
We demonstrate the strength of our proposed method quantitatively by evaluating its image retrieval performance on four public benchmark datasets. For a fair comparison to the previous arts~\cite{reality-check}, the accuracy of our model is computed in three different settings: we used 64, 128, and 512 embedding dimension on Cars-196, CUB-200-2011, and SOP datasets with the default image size $224 \times 224$.
On the In-Shop dataset the performance is only measured with embedding dimensions of 128 and 512 with the default image size $224 \times 224$.
Results on the Cars-196 and CUB-200-2011 datasets are exhibited in Table~\ref{t:cars_cub}. According to Table~\ref{t:cars_cub}, our model outperforms all the
previous state-of-the-art methods including the proxy-based~\cite{proxy-anchor,proxy-NCA,proxy-nca++}, pairwise-based~\cite{ranked-list,ms,softtriple} and ensemble methods~\cite{attention-ensemble} in all three settings often with a large margin on top 1 recall. In particular, on the challenging Cars-196 dataset, our method improves the previous best score by a large margin, 2.3$\%$, 5.3$\%$, and 2.4$\%$ in Recall$@1$ in embedding size of 64, 128, and 512 respectively. As reported in Table~\ref{t:SOP}, our model also achieves state-of-the-art performance on the SOP dataset. It outperforms previous methods in all cases except for Recall@10 and Recall@100 in 64-dimensional
embedding, but even in these cases, it achieves the second best. Finally, on the In-Shop dataset, it obtains the best scores in all two settings as shown in Table~\ref{t:Inshop}. On the In-Shop dataset, our model outperforms the state-of-the-art by a large margin of 2.7$\%$ in Recall@1.
In our experimental results, we noted that our model outperforms numerous state-of-the-art methods even with low-dimensional embedding vectors while they have higher embedding dimensions. This observation suggests that our model is capable of learning a more compact and effective embedding space.  Last but not least, our hybrid method converges in the same order as the proxy-based method as results are summarized in Figure~\ref{f:convergence}.
%
\begin{figure}[t]
    \centering
    \begin{tabular}{c}
  \includegraphics[width = 0.9 \linewidth]{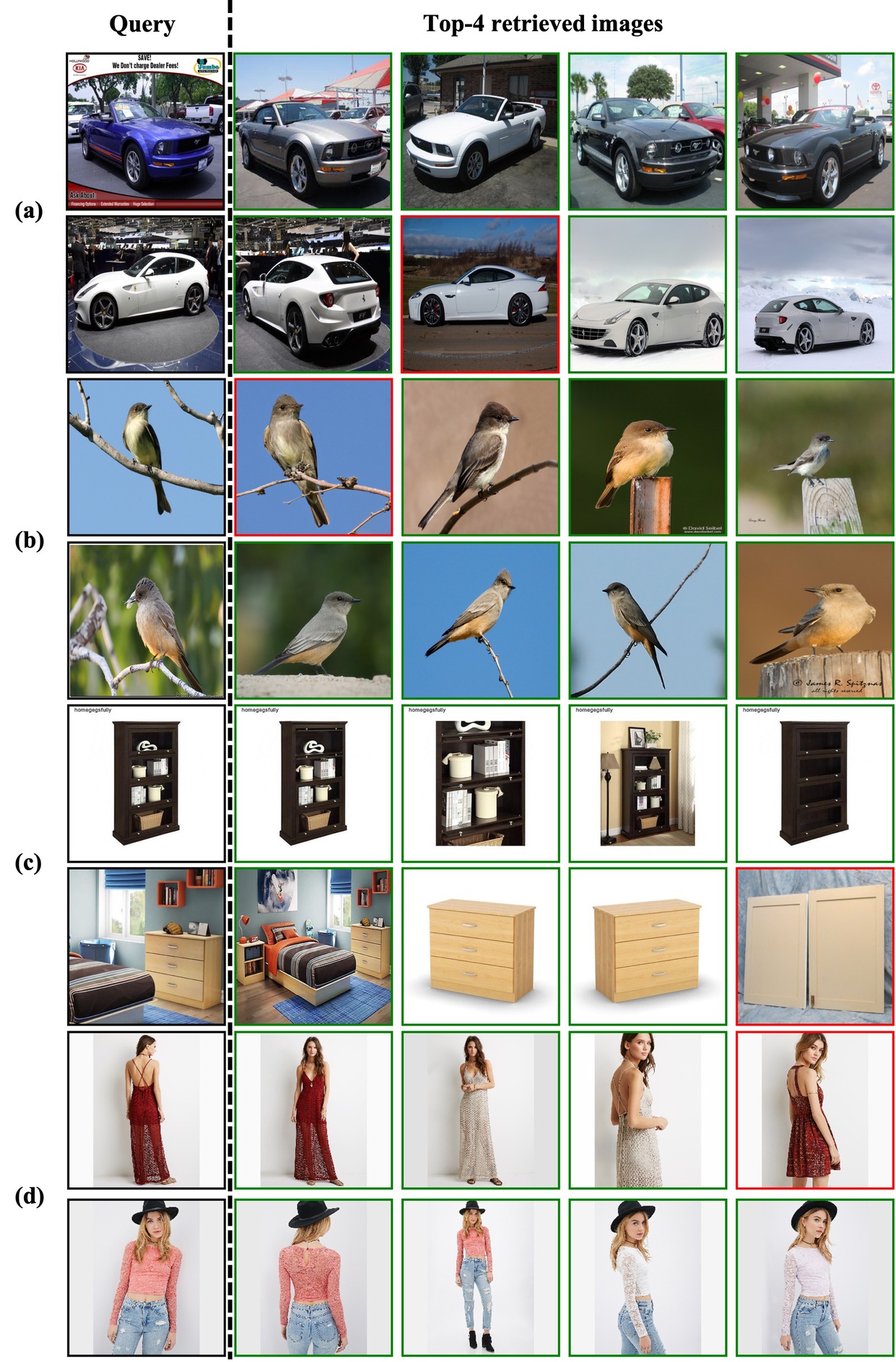}\\
    \end{tabular}
    \caption{Qualitative results on the Cars-196~(a) CUB-200-201l~(b), SOP~(c), and In-Shop~(d). For each query image (leftmost), top 4 retrievals are exhibited. The results with red boundaries are false cases but they are substantially similar to the query images in terms of appearance. (rows 2,3,6, and 7).}
    \label{f:qualitative_res}
\end{figure}
\begin{figure*}[t]
    \centering
    \begin{tabular}{ccc|cc}
    \multicolumn{3}{c}{(a)}&\multicolumn{2}{c}{(b)}\\
    & Cars-196& CUB-200-2011& Cars-196&CUB-200-2011 \\
  \rotatebox{90}{\hspace{6ex} R@1}&
  \includegraphics[width = 0.21 \linewidth]{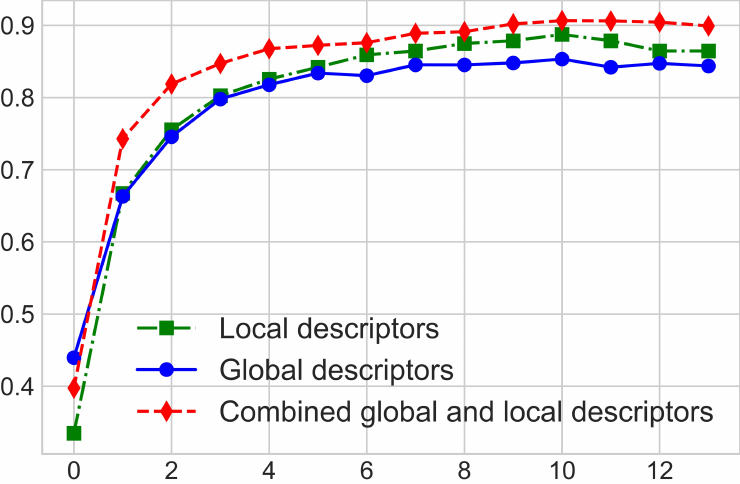} &\includegraphics[width = 0.21 \linewidth]{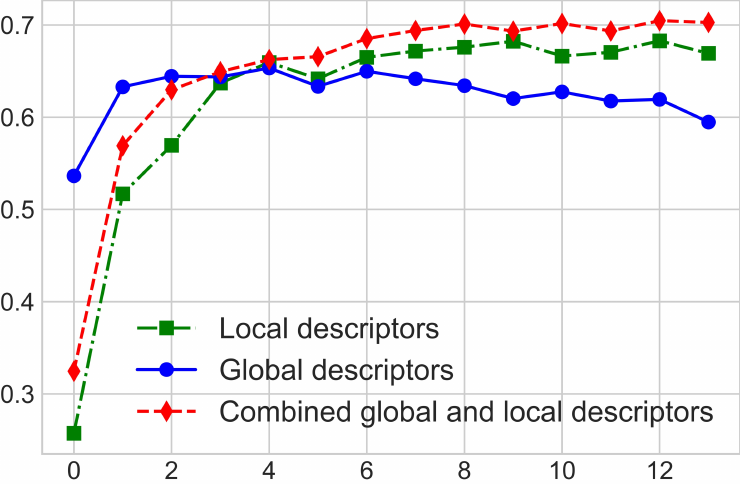}
  &\includegraphics[width = 0.21 \linewidth]{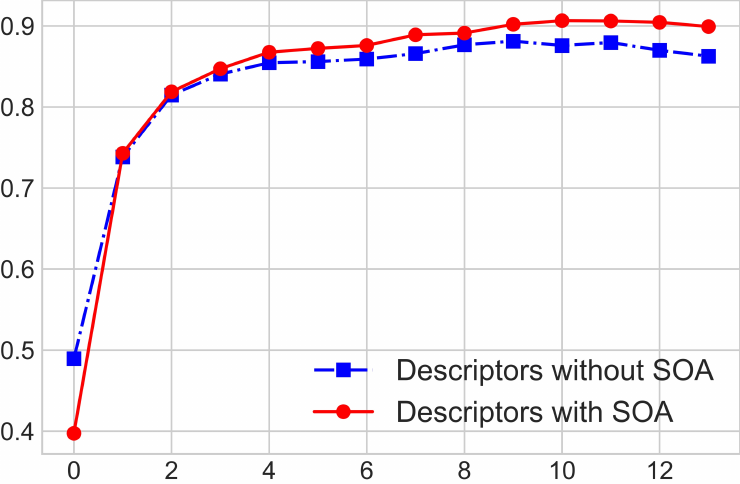} & \includegraphics[width = 0.21 \linewidth]{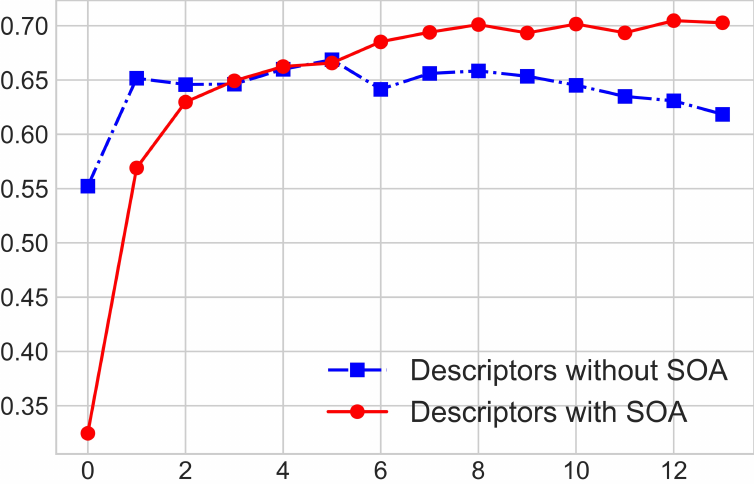}\\
  & Epochs &Epochs &Epochs &Epochs
    \end{tabular}
    \caption{(a): The impact of local and global descriptors in terms of Recall@1 performance on both Cars-196 and CUB-200-2011 datasets. The blue and green colors illustrate the performance of the global and local descriptors, while the red indicates the performance of the combined global and local descriptors. The~(a) part of this figure demonstrates the combination of local and global features are vital in our design. (b): The impact of the SOA component on the mentioned datasets. The blue color exhibits the model without second-order attention and red depicts the model with second-order attention.}
    \label{f:local_global_impacts}
\end{figure*}
\begin{figure*}[t]
    \centering
\scalebox{0.7}{
\begin{tabular}{c|c}
Cars196&CUB-200-2011 \\
\includegraphics[width = 0.65 \linewidth]{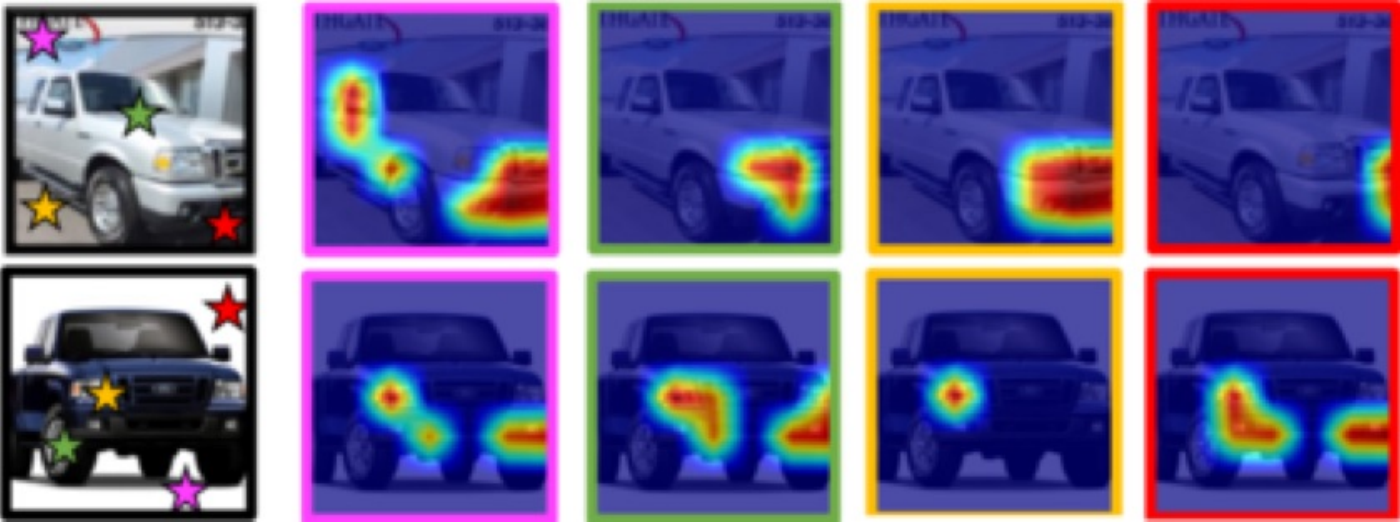}&
\includegraphics[width = 0.65 \linewidth]{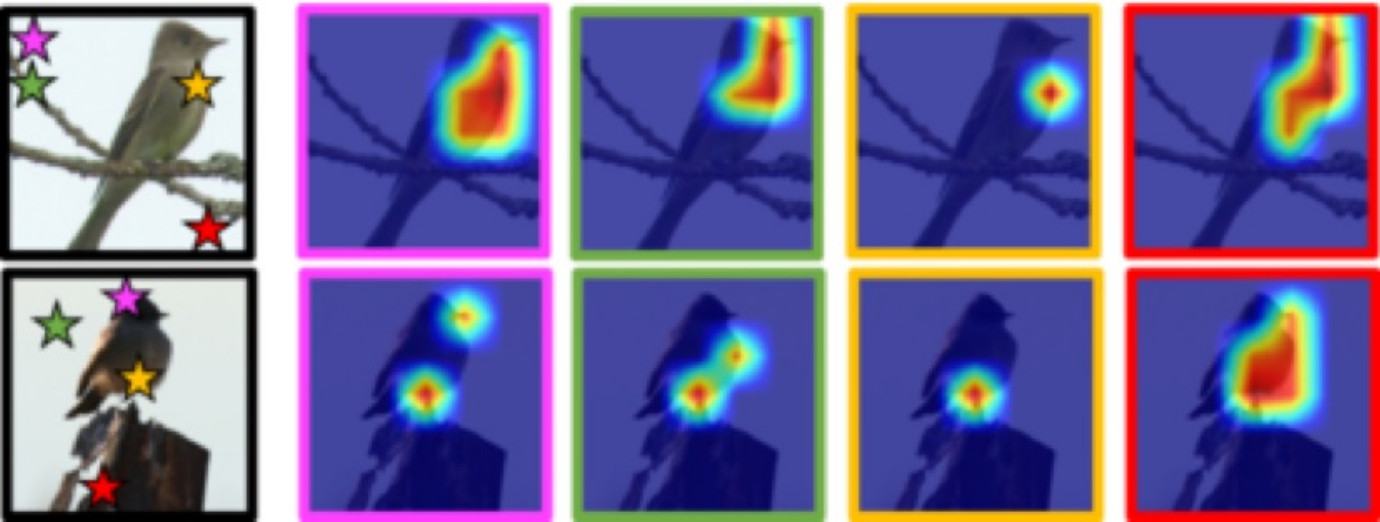}\\
\end{tabular}}
    \caption{Qualitative examples of SOA maps on the Cars-196 and CUB-200-2011 datasets. (a) corresponds to Cars196 dataset and (b) corresponds to CUB-200-2011 dataset. Each row depicts the source image and four corresponding SOA maps obtained for specific spatial locations (marked by stars).}
    \label{f:soa}
\end{figure*}
\subsection{Qualitative Results}
 \vspace{-0.2cm}
To further exhibiting the visual performance of our method, we
illustrate the qualitative retrieval results of our model on four datasets in Figure~\ref{f:qualitative_res}. Note that these datasets are challenging especially due to their large intra-class variations. For instance, the CUB200-2011 has a variety of poses and background clutter, the Cars-196 has various colors and shapes, and SOP and In-Shop datasets have challenging view-points of objects that make the retrieval tasks even harder. In contrast to all of these challenges, our proposed method performs robust retrieval.
\subsection{Ablation Study}
\vspace{-0.2cm}
\begin{figure}[t]
    \centering
    \begin{tabular}{ccc}
    & Cars-196& CUB-200-2011 \\
  \rotatebox{90}{\hspace{6ex} R@1}&
  \includegraphics[width = 0.4 \linewidth]{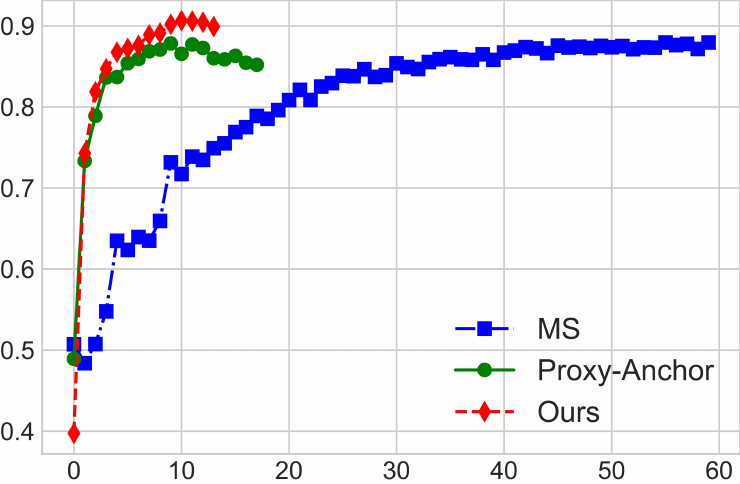} &\includegraphics[width = 0.4 \linewidth]{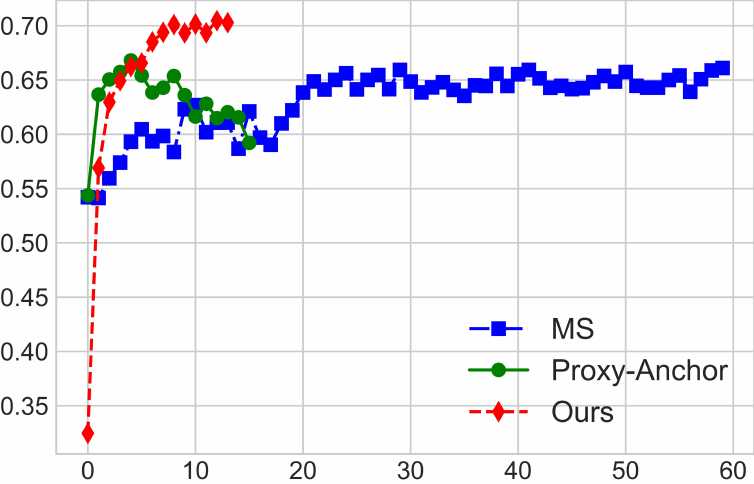}\\
  & Epochs &Epochs\\
    \end{tabular}
    \caption{The impact of the MS and Proxy-Anchor loss on Cars-196 and CUB-200-2011 datasets. The red color indicates our hybrid loss while the green and blue represent the Proxy-Anchor and MS losses, respectively. The Figure demonstrates the combination of two losses is crucial in our proposed method.}
    \label{f:loss_impact}
\end{figure}
\begin{figure}[t]
    \centering
    \begin{tabular}{cc}
  \rotatebox{90}{\hspace{6ex} R@1}&
  \includegraphics[width = 0.5 \linewidth]{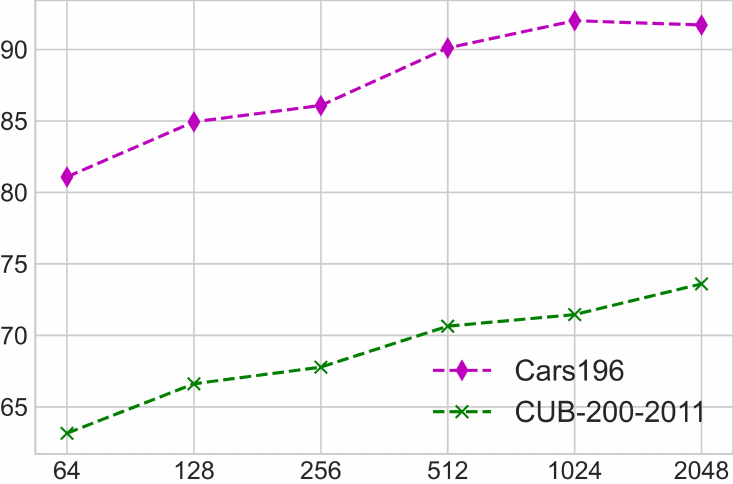} \\
  &Dimensions\\
    \end{tabular}
    \caption{Accuracy in terms of Recall@1 versus embedding dimension on both Cars-196 and CUB-200-2011 datasets.}
    \label{f:dim}
\end{figure}
\noindent\textbf{Local and Global Features.} To investigate the impact of the local and global features on the performance of our proposed method, we examined the Recall@1 while training our model with only local and global descriptors separately on the Cars-196 and CUB-200-2011 datasets. The result of the analysis is summarized in Figure~\ref{f:local_global_impacts} (a). This Figure illustrates the local descriptors on both Cars-196 and CUB-200-2011 datasets are getting slightly better performance (2$\%$) than the global descriptors but neither of these descriptors individually achieved the performance as the combination of these descriptors obtained on both datasets based on Recall@1 performance (see Table~\ref{t:cars_cub}). 

\vspace{0.18cm}
\noindent\textbf{Second-order Attention.} One of the vital components of our proposed method is SOA. We employ this attention mechanism to further enhance deep features based on their correlation. To evaluate the impact of the SOA, we trained our model with and without having the SOA, and the results are summarized in Figure~\ref{f:local_global_impacts} (b). According to this figure, the performance of the model significantly improves in terms of Recall@1 when the model is trained with SOA especially on critical datasets like CUB-200-2011. This observation confirms our assumption on the essence of higher-order attention for further enhancements of the representations.
We visualize the effects of second-order attention in Figure~\ref{f:soa}. This figure exhibits the attention map of samples from Cars-196 and CUB-200-2011 datasets. For each image, four parts have been selected with different stars and different colors. The attention map associated with each star has a border with identical colors. For locations in the background, interestingly, the attention from that feature is distributed within the main object in the image. It confirms the second-order attention has learned to focus on the main object in the image to accurately retrieve the most similar images to the query. On the other hand, when the star is located within the main object, the attention is on highly distinctive regions.

\vspace{0.18cm}
\noindent\textbf{A Single Head Attention.} Our proposed method requires two attention heads for local and global descriptors. We are interested in probing a test case with having only a single attentional head for both local and global descriptors. After extracting local descriptor $f_l \in \mathbb{R}^{14 \times 14 \times 1024}$ and global descriptor $f_g \in \mathbb{R}^{7 \times 7 \times 2048}$ from backbone, we use a focus layer~\cite{yolov4} to reshape the local descriptor $f_l^{new} \in \mathbb{R}^{7 \times 7 \times 4096} $ to spatially match the global descriptor. Then, we concatenate the global and local descriptors resulting in a combined feature map $f \in \mathbb{R}^{\in 7 \times 7 \times 6144}$. Then, we apply the SOA for further enhancement of this feature map. We evaluated this approach on Cars-196 and CUB-200-2011 datasets. The single head attention obtained 86.6$\%$ and 66.6$\%$ in terms of Recall@1 on Cars-196 and CUB-200-2011, respectively. Note the obtained results are slightly degraded from the multi-head approach 90.1$\%$ on Cars-196 and 70.6$\%$ on CUB-200-2011 (see Table~\ref{t:cars_cub}), that suggests the multi-head SOA is necessary.

\vspace{0.18cm}
\noindent\textbf{Multisimilarity and Proxy-anchor Loss.} The other crucial component of our architecture is the combination of pairwise-based and proxy-based loss functions. To study the impact of each loss function on our proposed method, we trained our model separately with MS loss and proxy anchor loss and we evaluated the performance based on Recall@1 on Cars-196 and CUB-200-2011 datasets. The results is exhibited in Figure~\ref{f:loss_impact}. This Figure demonstrates that the combination of these two losses are crucial in our design since none of them individually achieves our performance (see Table~\ref{t:cars_cub}).

\vspace{0.18cm}
\noindent\textbf{Embedding Dimension.} The dimension of embedding vectors is a vital factor that controls the trade-off between speed and accuracy in image retrieval systems. We thus investigate the effect of embedding dimensions on the retrieval accuracy in our method. We evaluated our model with embedding dimensions varying from 64 to 2048 following the experiment in~\cite{ms,proxy-anchor}. The result of the analysis is illustrated in Figure~\ref{f:dim}, in which the retrieval performance of our model is reported on both Cars-196 and CUB-200-2011 dataset. The performance of our loss is fairly stable when the dimension is equal to or larger than 128. The performance of our model on the Cars-196 dataset improves until reaching 1024 dimensional embedding and after that slightly degrades. On the other hand, the performance consistently increases with the embedding dimension on the CUB-200-2011 dataset showing that more information on that dataset helps the retrieval performance.
 \section{Conclusion}
 \vspace{-0.2cm}
 We have proposed a novel metric learning algorithm that takes the best of both proxy-based and pairwise-based losses. Also, it leverages enhanced local and global descriptors to improve the recall and precision simultaneously. Our method benefits from having a reach data-to-data relation as well as fast and reliable convergence. We extensively evaluated our model on 4 public benchmarks and our model has achieved state-of-the-art performance on all datasets in terms of Recall@1 accuracy. Also, our model converged quickly without any careful data sampling technique.
\section*{Acknowledgements}
\vspace{-0.2cm}
This research work was sponsored by the Office of Naval Research (ONR) via contract N00014-17-C-2007. Disclaimer: The views and conclusions contained herein are those of the authors and should not be interpreted as necessarily representing the official positions, policies or endorsements, either expressed or implied, of ONR or the U.S. Government.
{\small
\bibliographystyle{ieee_fullname}
\bibliography{egbib}
}
\newpage
\section{Additional Experimental Results}
This supplementary material exhibits additional experimental results excluded from the main paper due to space limitation. Section~\ref{s:batch_size} analyzes the impact of the batch size in our framework. Section~\ref{s:soa} visualizes attention maps generated from Cars-196~\cite{cars196} and CUB-200-2011~\cite{cub}. Section~\ref{s:qual_results} provides additional qualitative results including, image retrieval results of our model compared with MS~\cite{ms} and Proxy-Anchor~\cite{proxy-anchor} losses, and t-SNE visualization of the learned embedding space on four benchmark datasets~\cite{cars196,cub,sop,inshop}.
\subsection{Impact of Batch size}
\label{s:batch_size}
To investigate the impact of the batch size on the performance of our proposed method, we examined the Recall@1 by training our model on both Cars-196 and CUB-200-2011 datasets by varying batch sizes from 30 to 120. The result of the analysis is summarized in Table~\ref{t:batch_size} where it reveals our model is robust in terms of batch sizes, especially for batch sizes larger or equal to 60.
\subsection{Second-Order Attention}
\label{s:soa}
We visualize the effects of second-order attention in Figure~\ref{f:soa}. This Figure exhibits the attention map of samples from Cars-196 and CUB-200-2011 datasets. For each image, four parts have been selected with different stars and different colors. The attention map associated with each star has a border with identical colors. For locations in the background (row~2, col~2 or row~4, col~3), interestingly, the attention from that feature is distributed within the main object in the image. It confirms the second-order attention has learned to focus on the main object in the image to accurately retrieve the most similar images to the query. On the other hand, when the star is located within the main object, the attention is on highly distinctive regions(row~1, col~3 or row~2, col~4).
\begin{table}[t]
\caption{Recall@1 ($\%$) on the Cars-196 and CUB-200-2011 datasets with varying batch sizes from 30 to 120.}
\centering
\begin{tabular}{c|c|c}
&\multicolumn{2}{c}{Recall@1} \\
\hline
Batch Size & Cars-196 & CUB-200-2011\\
\hline
30&90.0&68.5\\
60&91.1 &70.6\\
90&91.7&70.7\\
120&91.2&70.8\\
\hline
\end{tabular}
\label{t:batch_size}
\end{table}
\subsection{Additional Qualitative Results}
\label{s:qual_results}
\begin{figure*}[t]
    \centering
\begin{tabular}{c}
\includegraphics[width = 0.9 \linewidth]{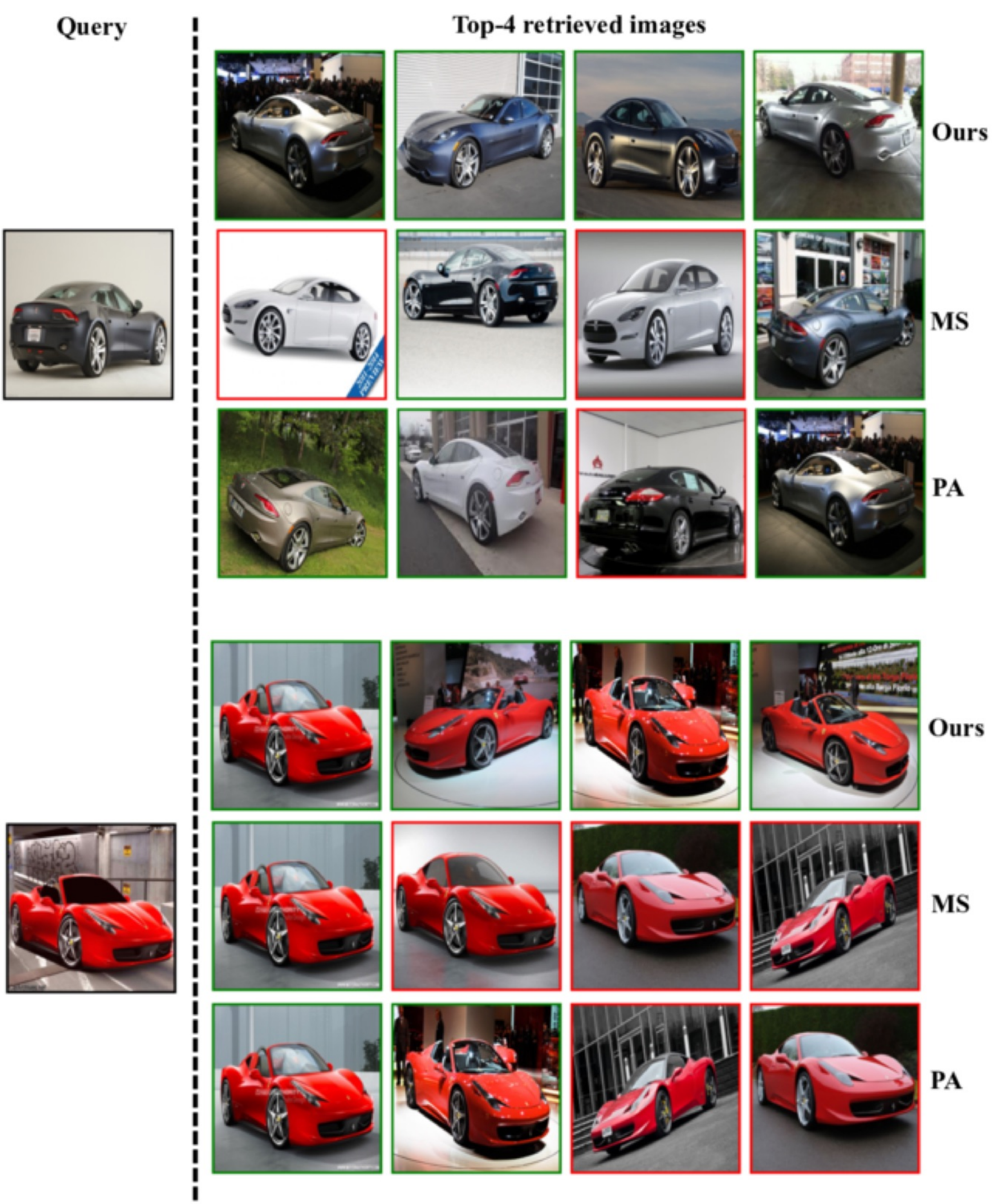} \\
\end{tabular}
    \caption{Qualitative results on the Cars-196 dataset generated from our hybrid loss, MS loss from the pairwise-based category, and Proxy-Anchor (PA) loss from proxy-based losses. For each query image (far most image), we retrieved the top 4 similar images.}
    \label{f:ret_cars}
\end{figure*}

\begin{figure*}[t]
    \centering
\begin{tabular}{c}
\includegraphics[width = 0.9 \linewidth]{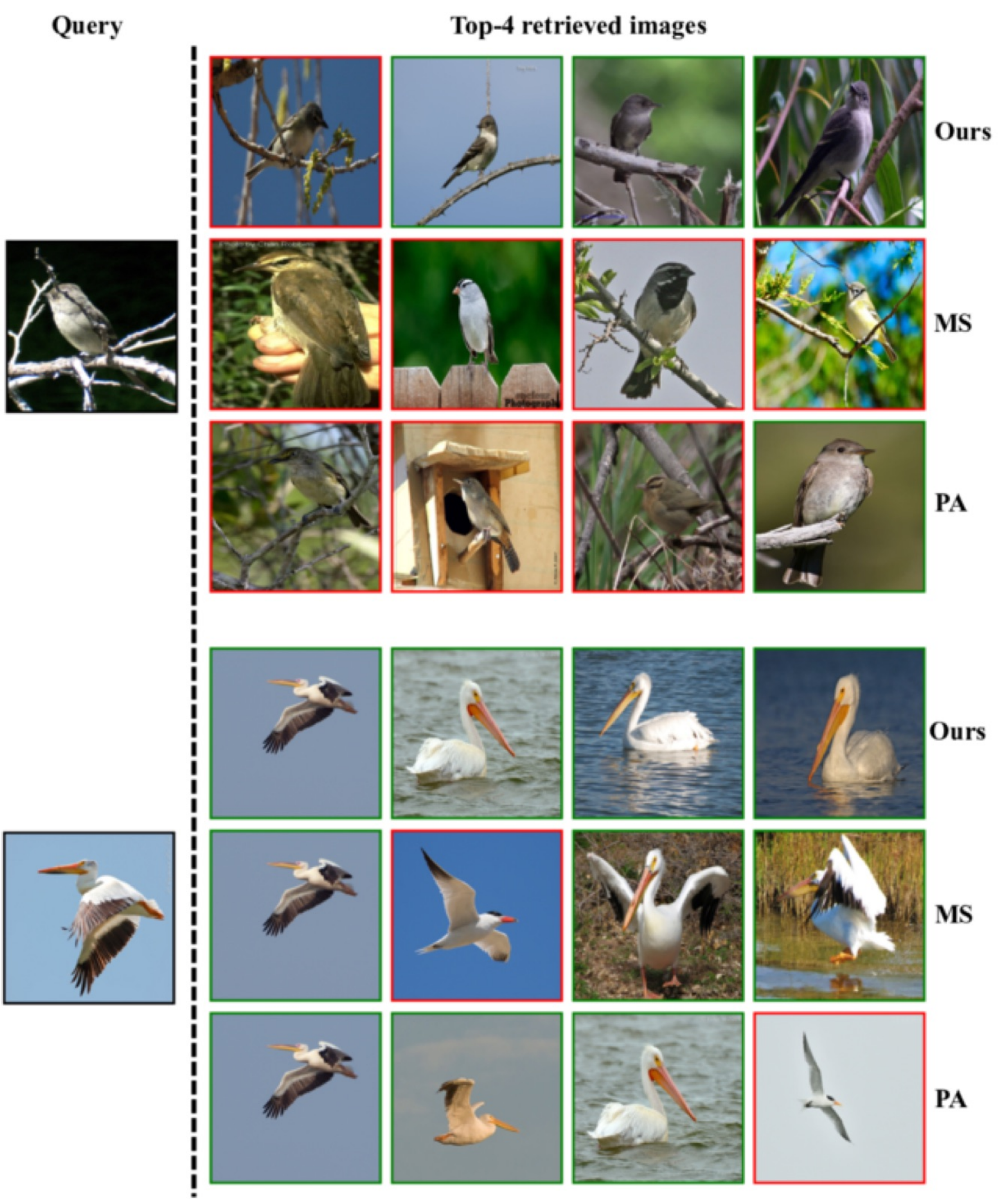} \\
\end{tabular}
    \caption{Qualitative results on the CUB-200-2011 dataset generated from our hybrid loss, MS loss from the pairwise-based category, and Proxy-Anchor (PA) loss from proxy-based losses. For each query image (far most image), we retrieved the top 4 similar images.}
    \label{f:ret_cub}
\end{figure*}

More qualitative examples for image retrieval on Cars-196 and CUB-200-2011 datasets are exhibited in Figure~\ref{f:ret_cars} and Figure~\ref{f:ret_cub}, respectively. The results of our model are compared with the MS loss --from the pairwise-based class-- and Proxy-Anchor --from the proxy-based category-- using the same backbone network. The overall results demonstrate that our model learned better embedding space than our baselines. In the examples in the 2nd, 3rd, 5th and 6th rows of Figure~\ref{f:ret_cars}, the MS loss and proxy anchor loss picked wrongful images while our hybrid loss retrieved the most similar cars to the query. Also, in Figure~\ref{f:ret_cub} MS and Proxy-Anchor loss missed several images, but our approach was able to retrieve the most similar images to the query. Even in Figure~\ref{f:ret_cub}, row 1 that our model wrongfully retrieved the first image, but as one can see, the image is significantly similar to the query image in terms of appearance. 

\begin{figure*}[t]
    \centering
\begin{tabular}{c}
\includegraphics[width = 0.95 \linewidth]{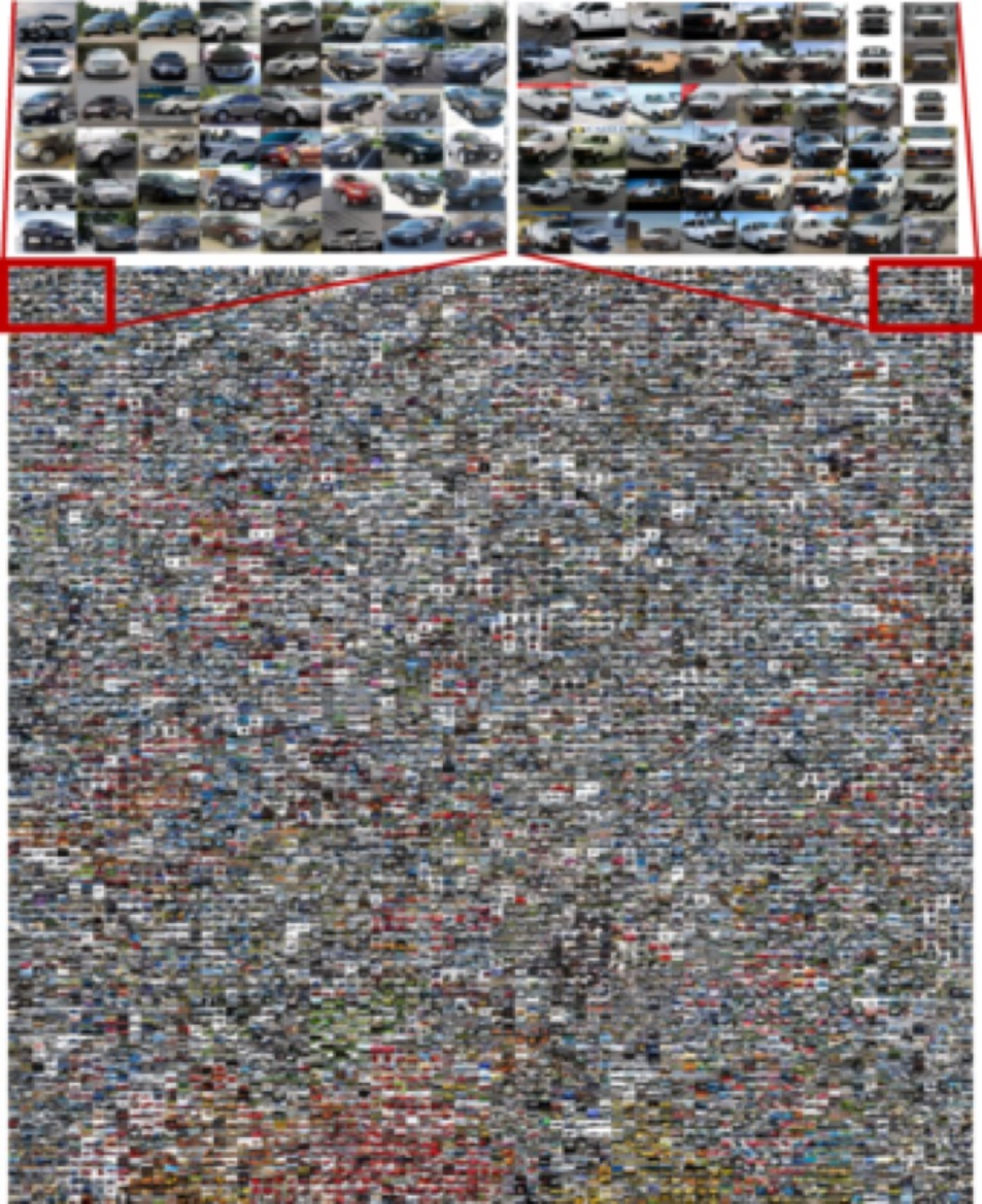} \\
\end{tabular}
    \caption{t-SNE visualization of our embedding space learned on the test set of Cars-196 dataset in a grid.}
    \label{f:tsne_cars}
\end{figure*}

\begin{figure*}[t]
    \centering
\begin{tabular}{c}
\includegraphics[width = 0.95 \linewidth]{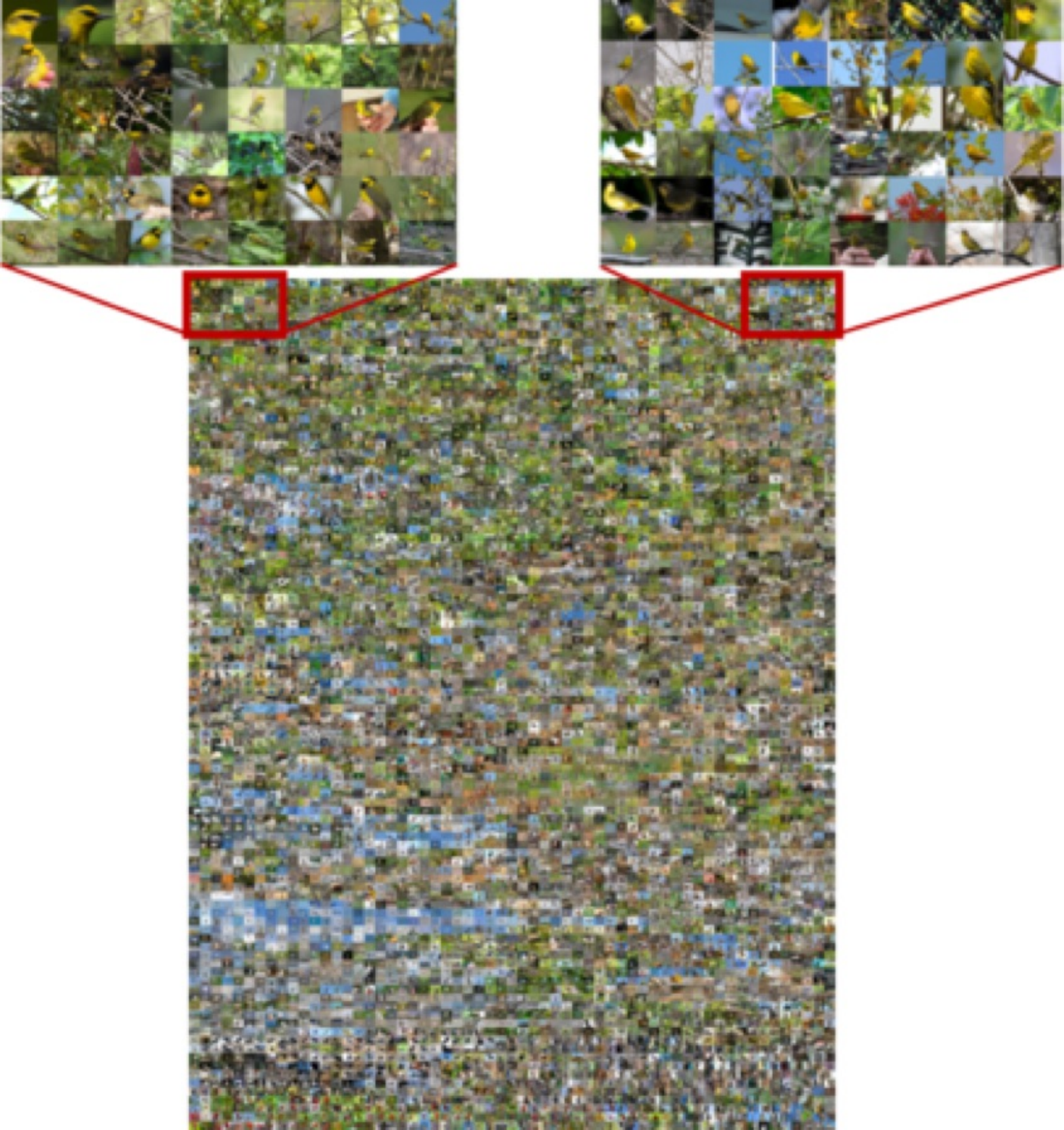} \\
\end{tabular}
    \caption{t-SNE visualization of our embedding space learned on the test set of CUB-200-2011 dataset in a grid.}
    \label{f:tsne_cub}
\end{figure*}

\begin{figure*}[t]
    \centering
\begin{tabular}{c}
\includegraphics[width = 0.95 \linewidth]{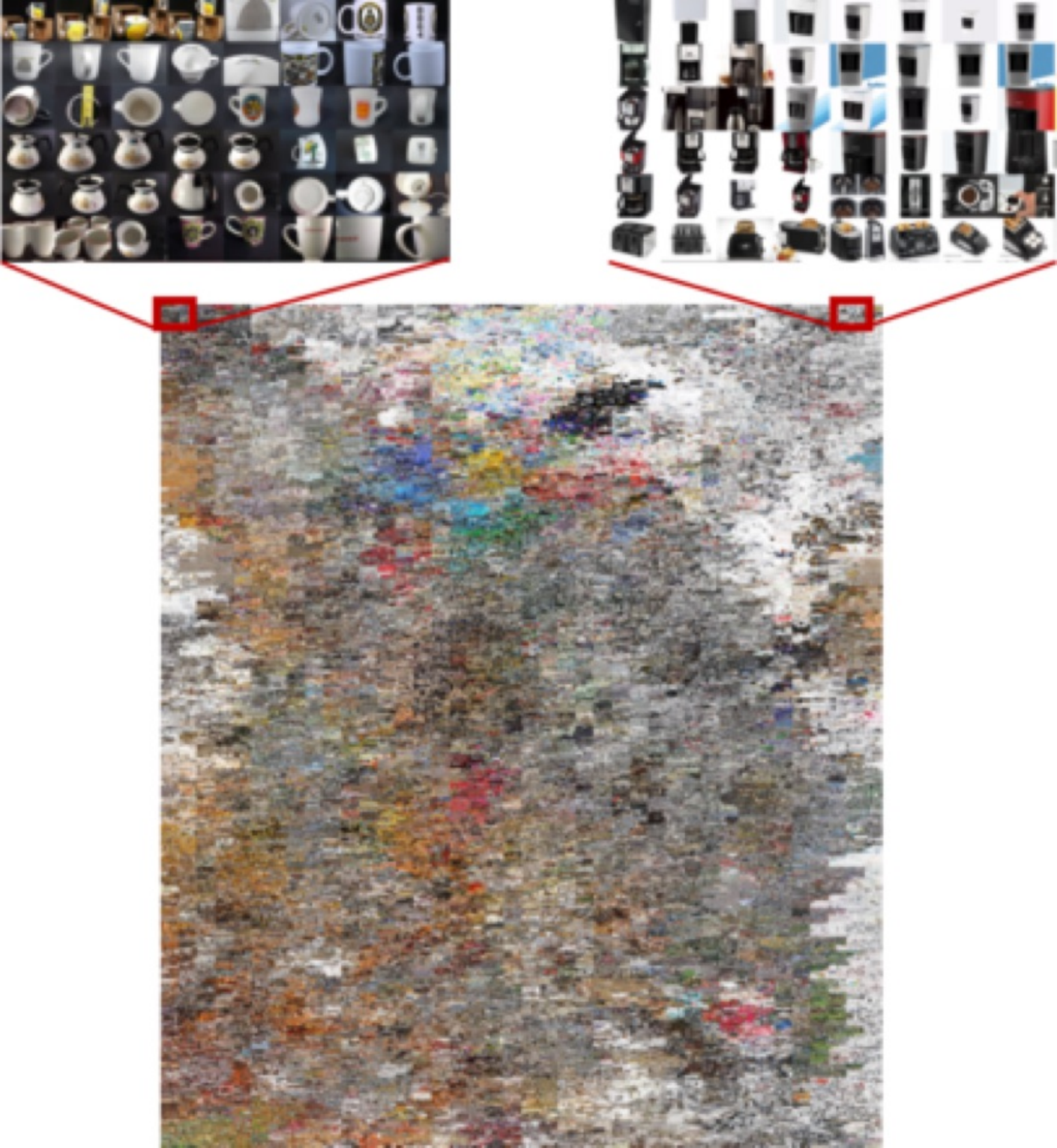} \\
\end{tabular}
    \caption{t-SNE visualization of our embedding space learned on the test set of SOP dataset in a grid.}
    \label{f:tsne_sop}
\end{figure*}

\begin{figure*}[t]
    \centering
\begin{tabular}{c}
\includegraphics[width = 0.90 \linewidth]{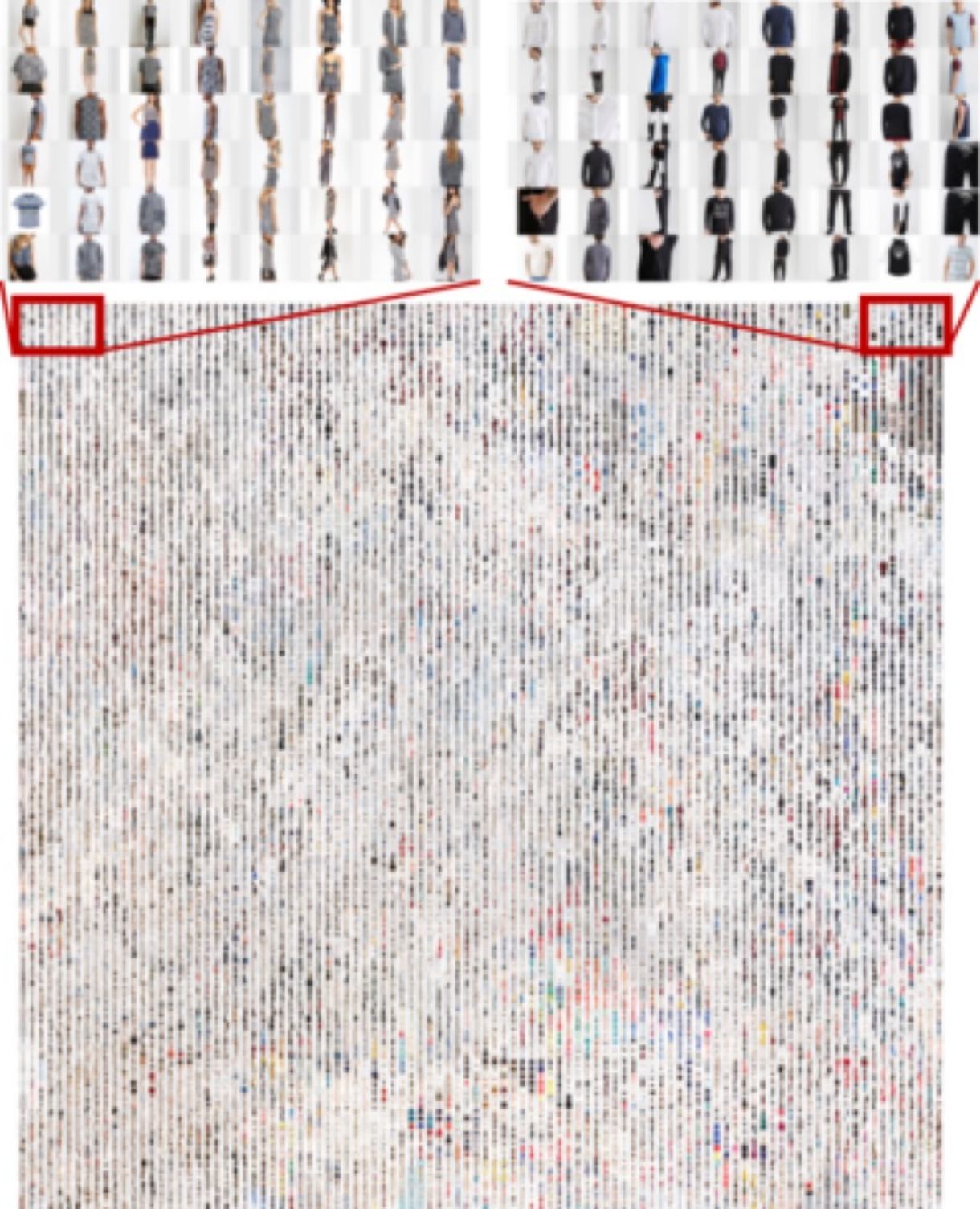} \\
\end{tabular}
    \caption{t-SNE visualization of our embedding space learned on the gallery split of Inshop dataset in a grid.}
    \label{f:tsne_inshop}
\end{figure*}

Based on the qualitative results given in Figures~\ref{f:ret_cars} and ~\ref{f:ret_cub}, one can see that our model produces accurate results. Also, the example in the first row of Figure~\ref{f:ret_cub} shows successful retrievals despite different view-point and color changes. Figures~\ref{f:tsne_cars},~\ref{f:tsne_cub} ,~\ref{f:tsne_sop} and~\ref{f:tsne_inshop} exhibit t-SNE visualizations of the embedding spaces learned by our hybrid loss on the test sets of the four benchmark datasets. The results demonstrate that all data points in the embedding space have relevant nearest neighbors, which suggests that our model learns a semantic similarity that can be generalized even in the test set.
\begin{figure*}[t]
    \centering
\begin{tabular}{c}
\includegraphics[width = 0.9 \linewidth]{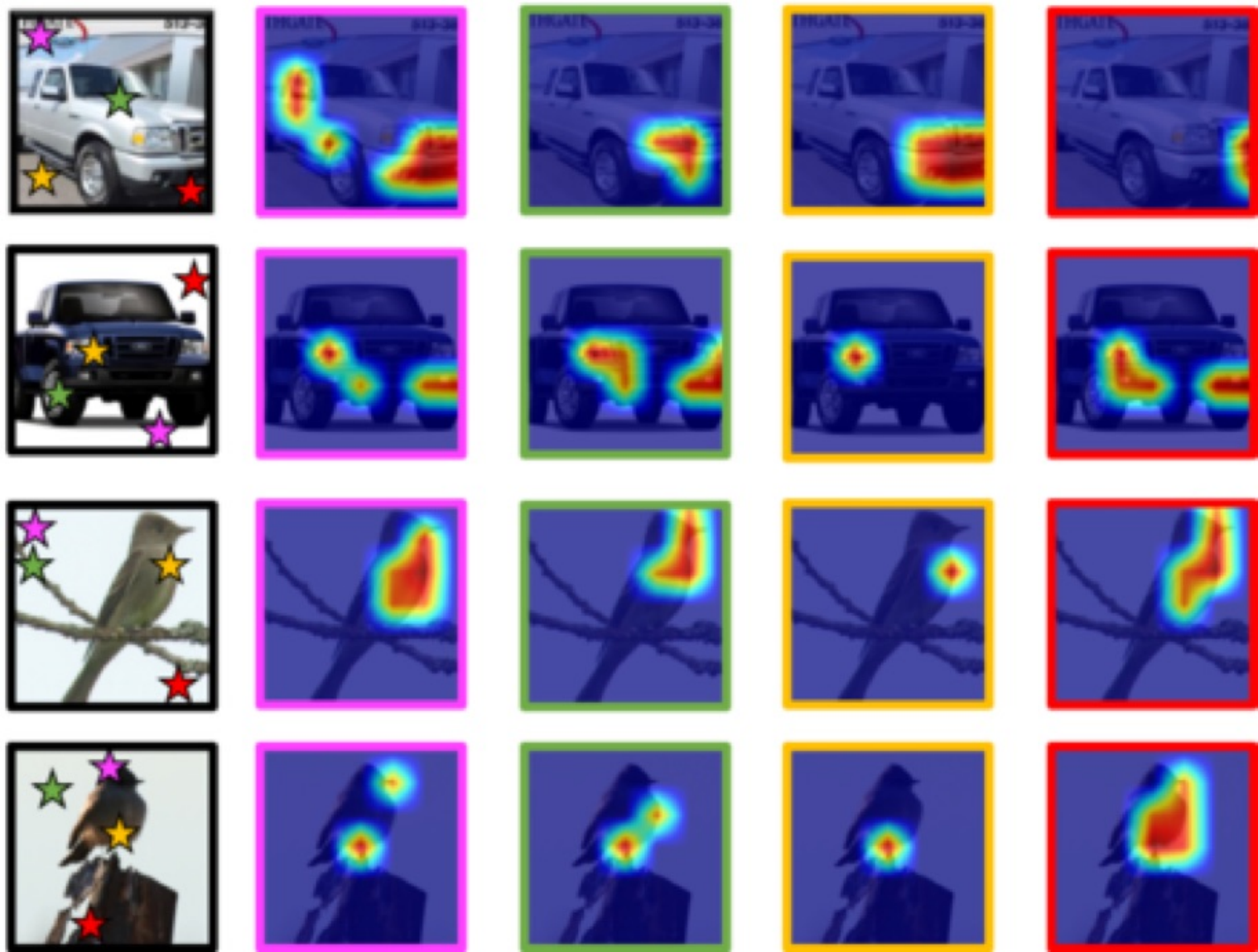} \\
\end{tabular}
    \caption{Qualitative examples of SOA maps on the Cars-196 and CUB-200-2011 datasets. Each row depicts the source image and four corresponding SOA maps obtained for specific spatial locations (marked by stars).}
    \label{f:soa}
\end{figure*}
\end{document}